\documentclass{article}

\usepackage{arxiv}

\usepackage[utf8]{inputenc} % allow utf-8 input
\usepackage[T1]{fontenc}    % use 8-bit T1 fonts
\usepackage{hyperref}       % hyperlinks
\usepackage{url}            % simple URL typesetting
\usepackage{booktabs}       % professional-quality tables
\usepackage{amsfonts}       % blackboard math symbols
\usepackage{nicefrac}       % compact symbols for 1/2, etc.
\usepackage{microtype}      % microtypography
\usepackage{lipsum}
\usepackage{graphicx}
\graphicspath{ {./images/} }
\usepackage[ruled,vlined,linesnumbered]{algorithm2e}

\usepackage{cite}
\usepackage{slashbox}
\usepackage{amsmath}
\usepackage{amssymb}
\usepackage{latexsym}
\usepackage{tabularx}
\usepackage{booktabs}
\usepackage{tabulary}

\title{Discriminative Domain-Invariant Adversarial Network for Deep Domain Generalization }

\author{
 Mohammad Mahfujur Rahman
 \\
  School of Electrical Engineering and Robotics\\
  Queensland University of Technology\\
 Queensland, Australia \\
  \texttt{m27.rahman@qut.edu.au} \\
   \And
 Clinton Fookes
 \\
  School of Electrical Engineering and Robotics\\
  Queensland University of Technology\\
 Queensland, Australia \\
  \texttt{
c.fookes@qut.edu.au} \\

  \And
Sridha Sridharan \\
School of Electrical Engineering and Robotics\\
Queensland University of Technology\\
Queensland, Australia \\
\texttt{s.sridharan@qut.edu.au} \\
}

\begin{document}
\maketitle
\begin{abstract}
Domain generalization approaches aim to learn a domain invariant prediction model for unknown target domains from multiple training source domains with different distributions. Significant efforts have recently been committed to broad domain generalization, which is a challenging and topical problem in machine learning and computer vision communities. Most previous domain generalization approaches assume that the conditional distribution across the domains remain the same across the source domains and learn a domain invariant model by minimizing the marginal distributions. However, the assumption of a stable conditional distribution of the training source domains does not really hold in practice. The hyperplane learned from the source domains will easily misclassify samples scattered at the boundary of clusters or far from their corresponding class centres. To address the above two drawbacks, we propose a discriminative domain-invariant adversarial network (DDIAN) for domain generalization. The discriminativeness of the features are guaranteed through a discriminative feature module and domain-invariant features are guaranteed through the global domain and local sub-domain alignment modules. Extensive experiments on several benchmarks show that DDIAN achieves better prediction on unseen target data during training compared to state-of-the-art domain generalization approaches.

\end{abstract}

%\footnotetext{Second footnote}
\makeatletter{\renewcommand*{\@makefnmark}{}
\footnotetext{ $^\star$ This manuscript is submitted to  Computer Vision and Image Understanding (CVIU)}\makeatother}

%\footnote{text}

% keywords can be removed

\section{Introduction}
\label{sec1}

\begin{figure*}
\begin{center}
\includegraphics[width=0.6\linewidth]{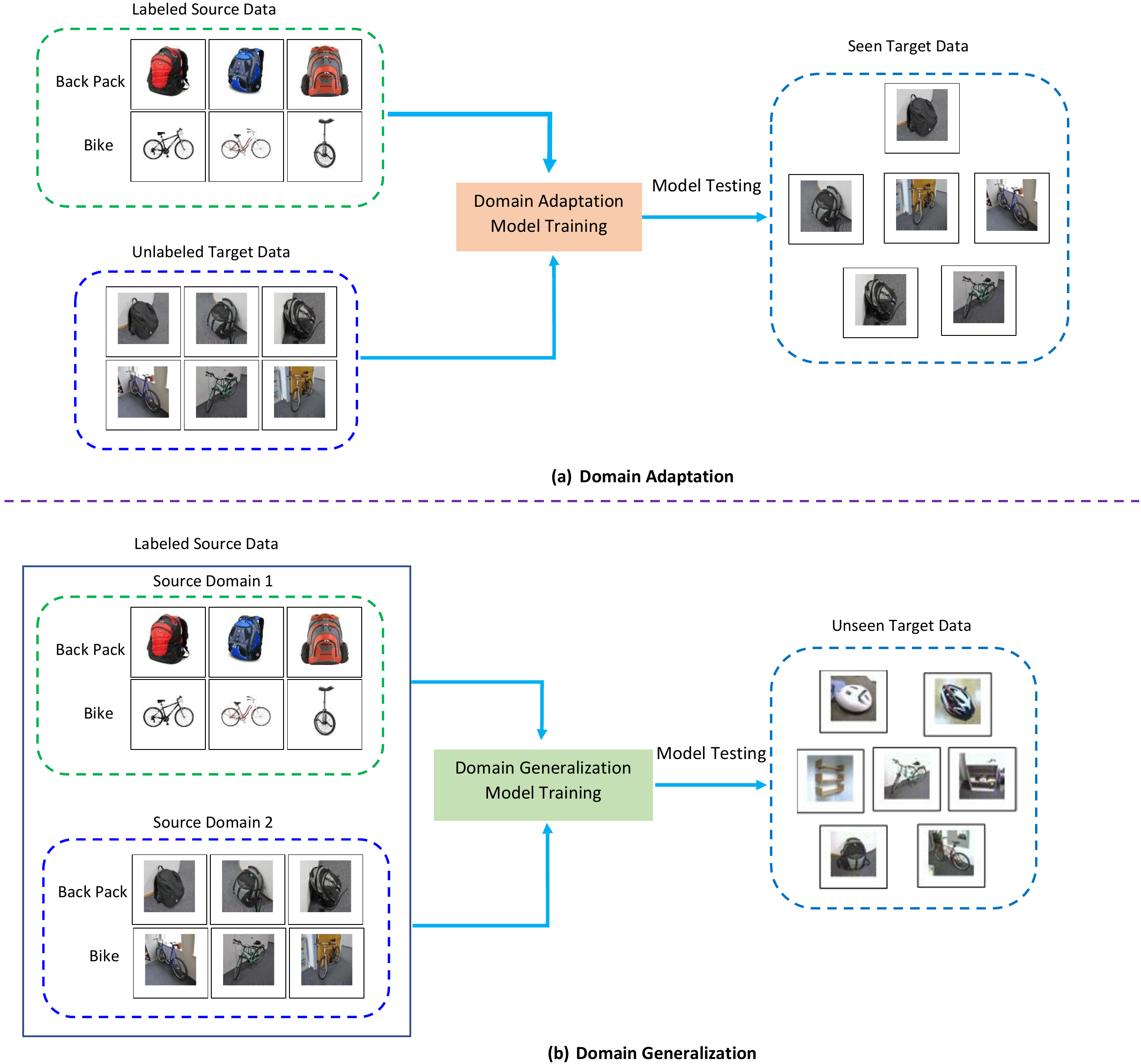}
\end{center}
   \caption{The comparison between the domain adaptation and domain generalization approaches. (a) Domain adaptation methods can access the unlabeled target data during training and the seen target data (seen in training phase) are evaluated in the test phase. (b) Domain generalization methods cannot access the target data during training and the unseen target data (unseen in training phase) are evaluated in the test phase.}
\label{fig:diff_DA_DG2}
\end{figure*}

Computer vision has attained remarkable progress with the developments in deep neural networks recently. Much of this progress has been achieved through the use of a supervised learning setting, which presumes that the training and testing samples follow an identical distribution. Nevertheless, this concept does not apply due to various shifting variables in many real-world situations, such as viewpoint changes, background noise and variance in lighting. These factors may induce bias in the collected datasets. Even powerful machine learning approaches such as deep learning may often decline quickly in performance due to dataset bias or if the training and test datasets have been collected from  non-identical distributions. To resolve these problems, domain adaptation \cite{chen2020adversarial,lee2019drop,Rahman2020,liu2019transferable,ma2019gcan,zhang2020unsupervised,mahfujur2021preserving,qiu2020partial,su2021multi} and domain generalization \cite{du2020learning,rahman2019multi,RAHMAN2020107124,li2018domain,qiao2020learning,ryu2019generalized,zhou2020learning,tseng2020cross,matsuura2020domain,chattopadhyay2020learning,seo2019learning,wang2020learning,mahfujur2021deep} approcehes have been proposed. Domain generalization is intended to handle the situation where there is no way to adapt to the target domain due to a lack of data. Compared to domain adaptation, domain generalization is a more challenging problem setting as explicit training on the target data is not allowed. Domain generalization learns domain-invariant feature representations from the given labeled data from multiple source domains and it generalizes well to unseen target domains without any further domain adaptation. Figure \ref{fig:diff_DA_DG2} shows the difference between domain adaptation and domain generalization.

Domain generalization is an active area of research which proposes a variety of approaches. Since there is no prior understanding of the target distribution, the crucial issue is how to lead the learning model to acquire discriminative representations for the particular task but is insensitive to domain-specific statistical shifts. Adversarial learning has recently been successfully integrated into deep networks to acquire transferable features to eliminate discrepancy in the distribution among the source domains to achieve domain invariant features that can be applied for the unseen target data. Recent advanced adversarial generalization methods \cite{Li2018eccv,Li_2018_CVPR,blanchard2017domain} demonstrated promising outcomes in various domain transfer tasks in the context of domain generalization.

The capacity to generalize to unknown environments is critical when machine learning models are applied to real-world conditions because the training and testing data come from different distributions. Domain generalization seeks to learn from multiple source domains a classification model and to generalize it to target domains that are not seen before. A critical problem involves learning domain-invariant representations in the generalization of domains. Significant efforts have recently been committed to broad domain generalization (DG). We are therefore proposing in this work a simple but effective model for the application of domain generalization to exploit adversarial learning to align both marginal and conditional distribution. 

Most previous research assumes that the conditional distribution among the source domains remains constant and that domain-invariant learning relies on the assurance of marginal distribution invariance. Most of these methods either align the global distributions across the source domains or align the conditional alignment. Li et al. \cite{Li2018eccv} suggested a conditional invariant approach for deep-domain generalization to optimise deep learning for seeking  domain-invariant features that uses class-specific domain identification mechanisms. Li et al. \cite{Li_2018_CVPR} utilize adversarial feature alignment via maximum mean discrepancy. Blanchard et al. \cite{blanchard2017domain} proposed a domain generalization method that predicts a classifier from the marginal distribution of the features. However, marginal and conditional distributions within domains sometimes lead to the adaptation in real applications differently. For instance, the marginal distribution is more important when the source domains are very dissimilar whereas the conditional distribution should be given more attention when the source domains are very similar. Most previous adversarial domain generalization approaches mostly adopt the discriminator which aligns the marginal distributions of the source domains. We introduce a framework for domain generalization that trains a shared embedding to align the marginal and conditional distributions and classes across the available source domains in order to obtain a domain agnostic model that can be applied for unseen target domains.

\begin{figure*}
\begin{center}
\includegraphics[width=1.0\linewidth]{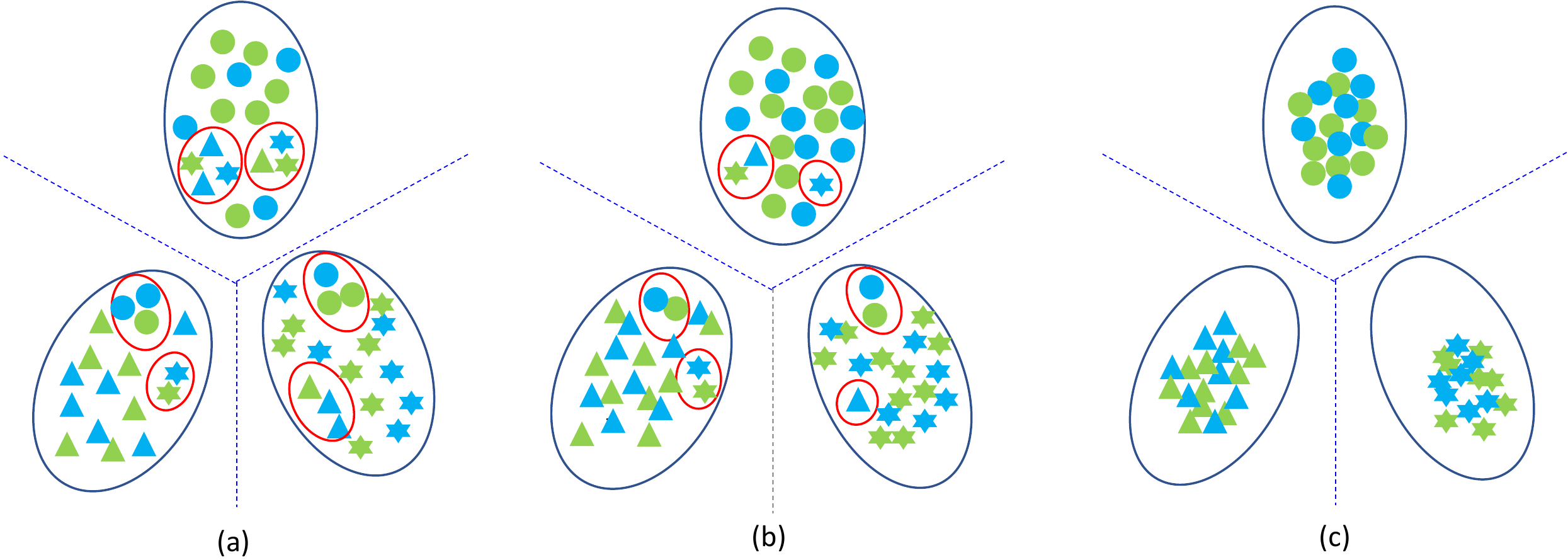}
\end{center}
 \caption{(Best viewed in color.) The importance of discriminative features learning for domain generalization. Green: samples from source domain 1; Blue: samples from source domain 2; Blue dot line: Hyperplane learned from the source domains during training; red circles indicate misalignment of the source samples; the Circle, Triangle and Star indicate three different categories, respectively. (a) Source Only (without using domain generalization), the hyperplane gained from the data of the source domains will be misaligned during training. (b) Domain alignment, the domain discrepancy has been reduced but not removed by the domain alignment. (c) Discriminative Feature Learning, the hyperplane gained from the data of the source domains can perfectly align the data according to their classes due to the discriminativeness of the domain invariant features.}
\label{fig:diff_DA_DG}
\end{figure*}

%Inspired from two more recent works \cite{wang2018visual,wang2017balanced} on domain adaptation align these two distributions, although they are based on a kernel approach with a higher computational cost. Moreover, \cite{wang2018visual} cannot handle large-scale data. 

%However, most of the existing work on domain generalization focuses only on learning to reflect common features by reducing the distribution disparity across domains. The domain alignment methods can only minimise but not eradicate the domain shift. The joint alignment ensures that not only the marginal distributions of the domains are aligned, but the conditional distributions as well. 

% Furthermore, to enforce the discriminative power of feature representations, with the labeled data in the source domains, we introduce the center loss. Extensive experiments and analysis on benchmark data sets demonstrate that the proposed approach achieves the best performance relative to state-of-the-art unsupervised approaches of domain generalization.

Moreover, most of the existing work on domain generalization focuses only on learning to reflect common features by reducing the distribution disparity across domains. The domain alignment methods can only minimise but not eradicate the domain shift. As a consequence, instances of a domain at the edge of clusters or far from their respective class centres are more probably to be incorrectly labeled by the hyperplane acquired from the source domains. A realistic way to alleviate this problem is to implement the samples with greater compactness in the intraclass. This will significantly reduce the number of samples that are far from the high density area and potentially misclassified. Likewise, by broadening the gap between different categories, another viable step is to remove the harmful effects of the domain disparity in aligned feature space. The effects of discrimnative features learning for domain generalization is shown in Figure \ref{fig:diff_DA_DG}. Fortunately, for the domain generalization task, the class information of the samples are available. In this regard, it is fair to render the source features more discriminatory in the matched feature space.

In this paper, we propose a novel Discriminative Domain-invariant Adversarial Network (DDIAN) for domain generalization. The proposed approach is capable of learning discriminative domain-invariant representations via end-to-end adversarial training. Stochastic Gradient Descent (SGD) will accomplish the adaptation with the gradients computed by backpropagation. The works's strengths include:

\begin{itemize}
    \item This is the first attempt, as far as we know, to jointly learn the deep discriminative feature and domain-invariant representations for deep domain generalization.
    
    \item Besides the marginal distribution, we also align the conditional distributions across the source domains.
    
   \item The experimental results prove that integrating the discriminative representation will further reduce the domain disparity and aid the ultimate classification task, which would greatly improve the performance of the generalization task. 
    
\end{itemize}

\section{Related Work}
\label{sec2}
Existing domain generalization methods for visual recognition can be divided into main two categories: shallow branch domain generalization and deep branch domain generalization.
%\end{enumerate}

\subsection{Shallow Branch Domain Generalization}

%The aim of domain generalization strives to decrease the domain shift issue similar to the problem of domain adaptation. As is formerly specified, the key distinction between domain adaptation and domain generalization is the unavailability of the unlabeled target data during the training phase. It has gained interest recently in classification tasks, particularly automatic gating of flow cytometry data \cite{NIPS2011_4312, Shi:2008:ATD:1431999.1432023} and object recognition \cite{FXR_iccv13, ECCV12_Khosla, Xu2014}. Therefore, domain adaptation approaches that can access unlabeled target data during training cannot be applied directly on domain generalization scenarios. 

Shallow branch domain generalization approaches are built as a two stage formulation. The first stage is utilized for extracting the features and the second stage is used for domain alignment. The issue of domain generalization was articulately commenced by \cite{NIPS2011_4312}. In \cite{NIPS2011_4312}, a kernel based classifier runs on multiple similar domains and the proposed approach is useful for solving automatic gating of flow cytometry. Specifically, \cite{NIPS2011_4312} adds all the training samples together in one dataset and it trains a single SVM classifier. Muandet et al. \cite{Muandet:2013:DGV:3042817.3042820} explored a kernel-based domain-invariant component analysis which is capable of learning a domain invariant transformation by decreasing the disparity across domains. It also preserves the operational correlation among the features and associated labels. A unified architecture for domain adaptation and domain generalization is proposed based on scatter component analysis in \cite{DBLP:journals/pami/GhifaryBKZ17}. Khosla et al. \cite{ECCV12_Khosla} proposed a multi-task max-margin classifier that measures the dataset-specific disparity in the feature space by adjusting the weights of the classifier. In \cite{DBLP:conf/iccv/GhifaryKZB15}, a multi-task autoencoder approach taking into account the construction capability of an autoencoder to extract domain-invariant features is introduced. Xu et al. \cite{Xu2014} add a nuclear norm-based regularizer which is capable of capturing the likelihoods of all positive samples to an exemplar-SVM for minimizing the domain discrepancy among source domains. Fang et al. \cite{FXR_iccv13} proposed unbiased metric learning by exploiting all information from the training source domains to train the classifier and produces a less biased distance metric that can be applied for object detection. Similarly, \cite{7733141} also exploit all the information from the training data to minimize the discrepancy across domains. In \cite{7298894} a robust classifier is learned reducing the domain bias among the training domains. These shallow domain generalization approaches require either hand crafted features or features extracted using pre-trained deep neural networks.

\subsection{Deep Branch Domain Generalization}

Deep branch domain generalization \cite{volpi2018generalizing,li2018domain,8237853,song2019episodic,balaji2018metareg,li2019feature,li2019feature,carlucci2019domain,shankar2018generalizing,asadi2019towards,deshmukh2019generalization,akuzawa2019adversarial,truong2018beyond,ilse2019diva,dou2019domain,blanchard2017domain,8053784,motiian2017CCSA,8237853,shankar2018generalizing,wang2019learning,li2019feature,RAHMAN2020107124,rahman2019multi} which is known as deep domain generalization incorporates the feature extraction and domain adaptation into a unified architecture. Deep domain generalization methods use a deep neural architecture to learn a model that can be applied to the target data. In these methods, the domain alignment module receives feedback from the feature extraction module and reinforces itself according to the feedback during training. Riccardo et al. \cite{volpi2018generalizing} proposed a deep domain generalization method based on adversarial data augmentation aiming to synthesize hard data at each iteration which are used to train the model to enhance its generalization capability. In \cite{li2018domain}, Maximum Mean Discrepancy (MMD) constraints are applied within the representation learning of an autoencoder via adversarial learning. Li et al. \cite{8237853} proposed an end-to-end low-rank parameterized convolutional neural network for domain generalization problem. In \cite{song2019episodic}, an episodic training attempts to learn a domain agnostic model by alternating domain-invariant feature extractors and classifiers among domains. Yogesh et al. \cite{balaji2018metareg} proposed a regularization function for the classification layer that can be helpful to apply for unknown target data in future. The classifier's weights are trained to achieve a more general classification model.

In \cite{li2019feature}, a feature-critic network is proposed that learns an auxiliary meta loss depending on output of the feature extractor. Carlucci et al. \cite{carlucci2019domain} solve domain generalization problem by jigsaw puzzles using maximal hamming distance algorithm. Shankar et al. \cite{shankar2018generalizing} generate domain-guided perturbations of input data that are utilised to train the model to obtain a robust model. \cite{DBLP:journals/corr/abs-1807-08479,li2018deep,motiian2017CCSA} use semantic alignment that attempts to make latent representation given class label identical within source domains. Recently, Akuzawa et al. \cite{akuzawa2019adversarial} proposed domain-invariant feature learning via adversarial learning as \cite{li2018deep}. \cite{akuzawa2019adversarial} needs one classifier whereas \cite{li2018deep} needs the same number of classifiers as the source domains. \cite{matsuura2019domain} combines multiple latent domains and train the model without using the domain label. It solves domain generalization problem using clustering strategies with adversarial learning.

\section{The Proposed Method}
In this section, we illustrate the proposed DDIAN architecture in detail that is displayed in Figure \ref{fig:arch_1}. The whole architecture consists of a feature extraction network, a classification network, a discriminative feature network, global domain alignment network and local sub-domain alignment network. Our aim is to achieve discriminative domain-invariant features that will aid the generalization of the domain. We will be explaining how we achieve this in the following section.
\label{sec2}

\subsection{Problem Definition}

Suppose \( \mathcal{X} \) represents the feature space and \( \mathcal{Y} \) represents the label space. A domain \( \mathcal{D} \) is denoted by a joint distribution $P(X, Y)$ defined on  \( \mathcal{X \times Y} \). We assume that we have a set of $N$ source domains such as $\Omega$ = $D_S^1$; $D_S^2$; $D_S^3$;\dots;$D_S^N$ and target domain $D_T^N$ where $D_T^N$ $\notin$ $\Omega$. The goal of DG is to gain a classifying function $f: X \rightarrow Y$ able to classify $\{x_i\}$ to the corresponding $\{y_i\}$ given $S_D^1$; $S_D^2$; $S_D^3$;\dots;$S_D^N$ during training as the input, but $T_D^N$ is unavailable during the training phase.

\subsection{Discriminative Domain-Invariant Adversarial Network (DDIAN)}
\subsubsection{Domain-invariant Feature Extraction}

\begin{figure*}
\begin{center}
\includegraphics[width=0.87\linewidth]{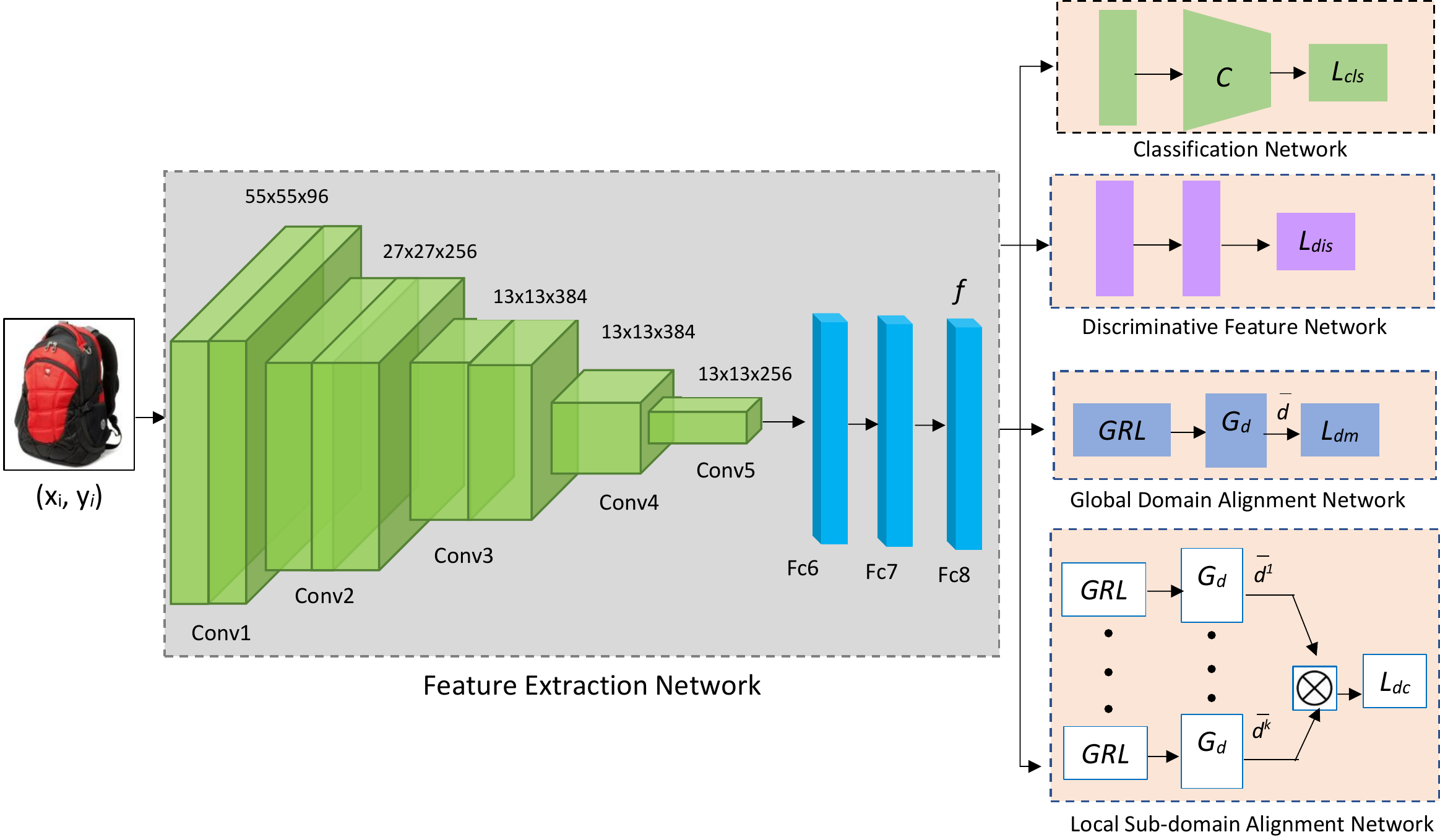}
\end{center}
   \caption{An overview of the proposed discriminative domain-invariant adversarial network. It consists of four sub-networks: feature extraction network, classification network, discriminative feature network and domain alignment network. Domain alignment network consists of two subnwtworks: global domain alignment network and local sub-domain alignment network. Either Alexnet or Resnet-18 pre-trained model is used to extract the features in the feature extraction network. Global and local sub-domain alignment networks are responsible to align the marginal and conditional distributions of the source domains during training. Discriminative feature network extracts discrminitive features which is helpful for final prediction by the classification network.}
\label{fig:arch_1}
\end{figure*}

Domain adversarial learning leverages the GAN concept to support transferable learning functionality. This learning technique is a two-players game. The name of the first player is domain discriminator $G_d$ which is trained to differentiate the source domains. On the other hand, the name of the second player is feature extractor $F$ which attempts to mislead the domain discriminator by retrieving domain invariant representations. These two players are adversarially trained: the parameters of the feature extractor ($\theta_f$) and discriminator ($\theta_d$) are learned by maximising and minimising the loss of the domain discriminator respectively. Moreover, the classifier $C$ loss is also decreased. We can formulate the loss function as,
\begin{equation}
L (\theta_f, \theta_c, \theta_d) = \frac{1}{n}\Bigg( \sum_{x_i \in \Delta} L_y \bigg(C(F(x_i)\bigg), y_i\Bigg) - \frac{\gamma}{n} \Bigg(\sum_{x_i \in \Delta} L_d \bigg(G_d(F(x_i)\bigg), d_i\Bigg),
\label{ch6:eq1}
\end{equation}
where $\gamma$ is a hyper-parameter, and $L_y$ and $L_d$ denote the classification loss and domain discriminator loss, $\theta_c$ is the classifier's parameters. $d_i$ denotes the domain label of the input instances. After the training converges, the parameters $\hat{\theta_f}$, $\hat{\theta_c}$ and $\hat{\theta_d}$ will deliver a saddle point of Eq. \ref{ch6:eq1},
\begin{align}
( \hat{\theta_f}, \hat{\theta_c}) = \underset{\theta_{f}, \theta_{c}}{\arg\min} L (\theta_{f}, \theta_{c}, \theta_{d})  \\
( \hat{\theta_d}) = \underset{\theta_{d}}{\arg\max} L (\theta_{f}, \theta_{c}, \theta_{d}).
\end{align}

Previous adversarial domain generalization methods either align the marginal distributions or align conditional distributions. The alignment of these two distributions has been shown to help improve the efficiency because both distributions are useful for acquiring domain-invariant representations. In this work, we extract the domain-invariant features using both the global domain alignment network and local sub-domain alignment network. The global domain alignment network is responsible for aligning the marginal distributions across the domains whereas the local sub-domain alignment network is responsible for aligning the conditional distributions among the domains.

We will first present the classification network, global domain alignment network, local sub-domain alignment network and discriminative feature network in the next sections. Then, we demonstrate the DDIAN loss function and procedure for training the DDIAN.

\subsubsection{Classification Network}

The category classifier ($C$, the green portion of Figure \ref{fig:arch_1}) is capable of classifying the categories of the input instances in the source domains. Hence the supervised or labeling information on the $x_i$ can be used. The training objective of the classifier is a cross-entropy loss that can be defined as,
\begin{equation}
L_{cls} = - \frac{1}{n} \Bigg(\sum_{x_i \in \Delta} \bigg(\sum^K_{c = 1} \hat{P_{x_i}} \log C(F(x_i))\bigg)\Bigg) ,
\label{ch6:eq4}
\end{equation}
where K is the number of classes of the source domains, $\hat{P_{x_i}}$ is the probability of the input sample $x_i$ belonging to category $K$, $F$ denotes the feature extractor and $C$ indicates the classifier.

\subsubsection{Global Domain Alignment Network}
The global domain discriminator is designed to eliminate the marginal distributions across the training source domains. The basic concept is to have the marginal domain discriminator accompanied by a domain-adversarial neural network \cite{Ganin:2016:DTN:2946645.2946704}. In \cite{Ganin:2016:DTN:2946645.2946704}, the domain discriminator is designed to align the marginal distributions between the source domain and target domain whereas we designed the global domain discriminator among the $n$ number of source domains. The loss of the marginal distribution which is achieved by the global domain discriminator can be formulated as,
\begin{equation}
L_{dm} = \frac{\gamma}{n} \Bigg(\sum_{x_i \in \Delta} L_d (G_d(F(x_i))), d_i\Bigg),
\label{ch6:eq5}
\end{equation}
where $L_d$ indicates the cross-entropy loss of the domain discriminator, $F$ depicts the feature extractor, and $d_i$ indicates the domain label of the input instances $x_i$.

\subsubsection{Local Sub-domain Alignment Network}

The discriminator in the local domain is structured to match the conditional distributions of the training source domains. The local domain discriminator is capable of coordinating the source distribution multi-mode structure with the global domain discriminator, allowing for more fine-grained domain adaptation. The local domain discriminator is divided into $K$ class-wise domain discriminators $G^K d$, each one responsible for matching the $K$ class-related details. The local-domain discriminator loss function can be formulated as,
\begin{equation}
L_{dc} = \frac{\beta}{n} \Bigg(\sum^K_{k=1} \bigg(\sum_{x_i \in \Delta} L^K_d (G^K_d(y^K_i, F(x_i))), d_i\bigg)\Bigg),
\label{ch6:eq6}
\end{equation}
where $L^K_d$ and $G^K_d$ indicate cross-entropy loss associated with class K and domain discriminator respectively. $y_i$ denotes the label of the input instances and $d_i$ indicates the domain label of the input instances of $x_i$.

\begin{algorithm*}[h]
\SetAlgoLined
\KwIn{Source labeled samples \{$X^s_i, Y^s_i$\} from source source domains $D_S^N$, and target unlabeled data  \{$X^t_i$\} from target domain $D_T^N$. Hyper-parameters $\alpha$, $\beta$ and $\gamma$}

\KwOut{Classifier $C$}

%\nl \bf Pass\;

  \For{iter from 1 to max-iter}{%
    Sample a mini-batch of source samples [$X^s_i, Y^s_i$] from source domains and target samples [$X^t_i$] from target domain \;
    \tcc{Update feature extraction, classification, discriminative feature, global domain alignment and and local sub-domain alignment network}  
    Compute $L_{cls}$ using $L_{cls} = - \frac{1}{n} \Bigg(\sum_{x_i \in \Delta} \bigg(\sum^K_{c = 1} \hat{P_{x_i}} \log C(F(x_i))\bigg)\Bigg)$. \\
    
    Compute $L_{dm}$ using $L_{dm} = \frac{\gamma}{n} \Bigg(\sum_{x_i \in \Delta} L_d (G_d(F(x_i))), d_i\Bigg)$. \\
    
    Compute $L_{dc}$ using $L_{dc} = \frac{\beta}{n} \Bigg(\sum^K_{k=1} \bigg(\sum_{x_i \in \Delta} L^K_d (G^K_d(y^K_i, F(x_i))), d_i\bigg)\Bigg)$. \\
    Compute $L_{dis}$ using $L_{dis} = \frac{1}{2} \sum^m_{i=1} \frac{\left\lVert F(x_i) - c_{y_i}\right\rVert^2_2} {\sum^K_{j=1, j \neq y_i} \left\lVert F(x_i) - c_{j}\right\rVert^2_2 + \phi}.$\\
    
    Update feature extraction, classification, discriminative feature, global domain alignment and and local sub-domain alignment network using $L = L_{cls} + \beta L_{dc} + \gamma L_{dm} + \alpha L_{dis}$.
    }
    \caption{{\bf Training procedure of DDIAN} \label{Algorithm:DDIA}}
\end{algorithm*}

\subsubsection{Discriminative Feature Network}

In order to enforce the feature extraction network to learn even more discriminative features, we introduce a center based discriminative representation learning method for domain generalization. It should be noted that the entire training process concentrates on the SGD mini-batch. Hence the discriminative loss mentioned below is also dependent on the batch of instances. Since, the labels of the training samples are available for the source domain, the features of the source domains will be classified by the classifier. Furthermore, it is important to keep the discriminative power of feature representations during domain alignment. Although the distributions of the source domains are aligned, there may still be some samples falling into inter-class gaps, which proposes the requirement for learning more discriminative features.

There exists several methods for learning discriminative features \cite{khatun2021end,khatun2020joint,khatun2018deep,khatun2020semantic}, such as the triplet loss, the contrastive loss and the center loss. Both the triplet loss and the contrastive loss need to construct a lot of image pairs and compute the distance between images of each pair, which is computationally complicated. Therefore, in this study, we introduce the center loss, which can be flexibly combined with the above classification loss. The features derived from the deep neural network trained under softmax loss supervision are separable, but not as discriminatory because they show significant variations in intra-class distance. In \cite{wen2016discriminative}, authors build an efficient loss function based on the hypothesis to increase the power of the deep features taken from deep neural networks. Center loss mitigates the intra-class distances, on the other hand the softmax loss is used to classify the features corresponding to their categories. Influenced by the center loss which penalises the distance of each sample to the corresponding class centre, we proposed the discriminative feature learning as below,
\begin{equation}
L_{c} = \frac{1}{2} \sum^m_{i=1} \left\lVert F(x_i) - c_{y_i}\right\rVert^2_2,
\label{ch6:eq7}
\end{equation}
where $L_c$ indicates the center loss, m indicates the number of the training instances in a mini-batch, $x_i \in R_d$ indicates the $i$th training instances, $y_i$ indicates the label of $x_i$. $c_{y_i} \in R_d$ indicates the $y_i$th class of deep features and d indicates the deep feature dimension.

Discriminative representations should have greater separability within groups and intra-class compactness. Center loss utilizes the Equation \ref{ch6:eq7} loss function to penalise large distances in the intra-class. Nevertheless, the lack of center loss is that it does not acknowledge the separability of the inter-class. As we know, if the distances of the different classes are far enough, the representations will be more discriminative for the greater separability of classes. Therefore, because the centre loss just penalises broad intra-class distances, and does not include inter-class distances, the inter-class adjustment is minimal, ensuring the class centre positions can change slightly throughout the training process. As a consequence, if the network initialises the class centres using a relatively smaller variance, the smaller differences between the class centres would lead during training as the centre loss just penalises the wide intra-class distances without taking into account the inter-class distances. The center loss vulnerability is that it does not acknowledge the separability of the inter-class.

We are therefore proposing a new loss function to acknowledge inter-class separability and intra-class compactness concurrently by penalising the the sum of the distances of training samples to their non-related class centres and contrasting values between the distances of training data to their respective class centres as,
\begin{equation}
L_{dis} = \frac{1}{2} \sum^m_{i=1} \frac{\left\lVert F(x_i) - c_{y_i}\right\rVert^2_2} {\sum^K_{j=1, j \neq y_i} \left\lVert F(x_i) - c_{j}\right\rVert^2_2 + \phi},
\label{ch6:eq8}
\end{equation}

where $L_{dis}$ denotes the discriminative loss. $m$ denotes the number of training samples in a mini-batch. $F(x_i) \in R_d $ denotes the deep features of the $i$th training sample with dimension $d$.

\subsubsection{Overall Objective}

The overall objective of the model can be formulated as:
\begin{equation}
L = L_{cls} + \beta L_{dc} + \gamma L_{dm} + \alpha L_{dis},
\label{ch6:eq9}
\end{equation}
where $\gamma, \beta$ and $\alpha $ are weighted parameters. $L_{cls}$ is the classification loss, $L_{dm}$ is the marginal adversarial loss, $L_{dc}$ is the conditional adversarial loss and $L_{dis}$ is the discriminative loss. Algorithm \ref{Algorithm:DDIA} describes the overall training procedure of our proposed method.

\section{Evaluation and Testing}

\begin{table*}[h!]
%\fontsize{10}{10}\selectfont 
%\fontsize{7}{7}\selectfont 

\small\addtolength{\tabcolsep}{10.5pt}
\centering
 %\begin{tabular}{||c c c c||} 
 \resizebox{14.5cm}{!}{
 %\resizebox{\columnwidth}{!}{

  \begin{tabular}{|c | c  c c c c|} 
 \hline
 Source $\rightarrow$ Target  &P, C, S $\rightarrow$ A &P, A, S $\rightarrow$ C & A, C, S $\rightarrow$ P & A, C, P $\rightarrow$ S  &Ave. \\ [0.5ex] 
 %\hline\hline
  \hline
  
 % SCA \cite{DBLP:journals/pami/GhifaryBKZ17} &50.05 &58.79 &59.10 &50.62 &54.64 \\
  
 % MTAE \cite{DBLP:conf/iccv/GhifaryKZB15} &45.95 &51.11 &58.44 &49.25 &51.19 \\
  
Source only &64.9 &64.3 &86.7 &55.1 &67.2 \\
%ERM &63.2 &66.7 &86.6 &61.6 &69.5 \\

MASF &70.4 &72.5 &90.7 &67.3 &75.2 \\

CIDDG \cite{Li_2018_ECCV} &62.7 &69.7 &78.7 &64.5 &68.9 \\

DBADG \cite{8237853}  &62.9 &67.0 &\textbf{89.5} &57.5 &69.2 \\

DSN \cite{NIPS2016_6254} &61.1 &66.5 &83.3 &58.6 &67.4 \\

MLDG \cite{li2018learning} &66.2 &66.9 &88.0&59.0 &70.0 \\

CrossGrad \cite{shankar2018generalizing} &61.0 &67.2 &87.6 &55.9 &70.0 \\

MetaReg \cite{balaji2018metareg}  &63.5 &69.5 &87.4 &59.1 &69.9 \\

D-SAM \cite{d2018domain} &63.9 &70.7 &64.7 &85.6 &71.2 \\

JiGen \cite{carlucci2019domain} &67.6 &71.7 &89.0 &65.2 &73.4 \\

 Epi-FCR \cite{song2019episodic} &64.7 &\textbf{72.3} &86.1 &65.0 &72.0 \\
  
 % Metareg &69.82 &70.35 &91.0 &59.26 & \\
   \hline
 \textbf{DDIAN (Ours)} &\textbf{67.8} &69.6 &89.2 &\textbf{66.2} &\textbf{73.2} \\
  
 % \hline
 %Ave.  & & & & &\\ [1ex] 
 
 \hline
 \end{tabular}}
  \caption{Recognition accuracies for DG on the PACS dataset \cite{8237853} using pretrained AlexNet.}
\label{table_pacs_AlexNet}
\end{table*}

\begin{table*}[h!]
\fontsize{7}{7}\selectfont  
\centering
\small\addtolength{\tabcolsep}{11.5pt}
 %\begin{tabular}{||c c c c||} 
  \resizebox{14.5cm}{!}{
  \begin{tabular}{|c |  c  c c c c|} 
 \hline
 Source $\rightarrow$ Target  &P, C, S $\rightarrow$ A &P, A, S $\rightarrow$ C & A, C, S $\rightarrow$ P & A, C, P $\rightarrow$ S  &Ave. \\ [0.5ex] 
 %\hline\hline
  \hline
  
  Source only &77.8 &74.3 &94.7 &69.1 &79.0 \\ 
  
  %CIDDG \cite{Li_2018_ECCV} & & & & & \\
  
  %DBADG & & & & & \\
  
  %DSN &85.58 &77.29 &93.59 &72.00 & \\
  %ERM &75.8 &77.8 &95.4 &69.8 &79.7 \\
    MASF &80.3 & 77.2 &95.0 &71.7 &81.0 \\
  
  MLDG \cite{li2018learning} &79.5 &77.3 &94.3 &71.5 &80.7 \\
  
  MAML \cite{finn2017model}  &78.3 &76.5 &95.1 &72.6 &80.6 \\
  
  CrossGrad \cite{shankar2018generalizing} &78.7 &73.3 &94.0 &65.1 &77.8 \\
  
  MetaReg \cite{balaji2018metareg} &79.5 &75.4 &94.3 &72.2 &80.4 \\
 
  D-SAM \cite{d2018domain} &77.3 &72.4 &77.8 &95.3 &80.7 \\
  
  JiGen \cite{carlucci2019domain} &79.4 &75.3 &96.0 &71.4 &80.5 \\
  
  Epi-FCR \cite{song2019episodic} &82.1 &\textbf{77.0} &93.9 &73.0 &81.5 \\
  
 % Metareg &87.2 &79.2 &97.6 &70.3 \\
  \hline
 % \textbf{FNN(Ours)} &86.71 &78.49 &95.38 &74.94 &83.88 \\
  
   % \textbf{CFNN(Ours)} &\textbf{88.26} &\textbf{81.10} &\textbf{96.40} &\textbf{76.55} &\textbf{85.59}  \\
  \textbf{DDIAN (Ours)} &\textbf{83.4} &76.7 &\textbf{95.3} &\textbf{74.1} &\textbf{82.4} \\
  
 % \hline
 %Ave.  & & & & &\\ [1ex] 
 
 \hline
 \end{tabular}}
  \caption{Recognition accuracies for DG on the PACS dataset \cite{8237853} using pretrained ResNet-18.}
\label{table_pacs_ResNet}
\end{table*}

\begin{table*}[h!]
%\fontsize{7}{7}\selectfont 
\centering
\fontsize{7}{7}\selectfont  
\centering
\small\addtolength{\tabcolsep}{11.5pt}
 %\begin{tabular}{||c c c c||} 
  \resizebox{14.5cm}{!}{
 \begin{tabular}{|c |  c c c c c |} 

 \hline
 Source $\rightarrow$ Target  &L, P, S $\rightarrow$ C &P, C, S$\rightarrow$L &C, L, S$\rightarrow$P & P, L, C$\rightarrow$S  &Ave. \\ [0.5ex] 
 %\hline\hline
  \hline
  %1NN &87.65 &67.00 &97.36 &82.11 &83.00 \\

  Source only   &85.7 &61.3 &62.7 &59.3 &67.3 \\

  CIDG \cite{DBLP:journals/corr/abs-1807-08479}  &88.8 & 63.1&64.4 &62.1 &69.6 \\

 CCSA  &92.3 &62.1 &67.1 &59.1 &70.2 \\
 
 SLRC &92.8 &62.3 &65.3 &63.5 &71.0 \\
 
 DBADG &93.6 &63.5 &70.0 &61.3 &72.1 \\

 MMD-AAE &94.4 &62.6 &67.7 &64.4 &72.3 \\
 
 D-SAM &91.8 &57.0 &58.6 &60.9 &67.0\\
 JiGen &96.9 &60.9 &70.6 &64.3 &73.2 \\
  \hline
  \textbf{DDIAN(Ours)} &95.7 &\textbf{64.8} & 69.2 &\textbf{65.1} &\textbf{73.7} \\
 
 \hline
 \end{tabular}}
  \caption{Recognition accuracies for DG on the VLCS dataset  using pretrained AlexNet.}
\label{vlcs}
\end{table*}

\begin{table*}[h!]
\fontsize{7}{7}\selectfont  
\centering
\small\addtolength{\tabcolsep}{11.5pt}
 %\begin{tabular}{||c c c c||} 
  \resizebox{14.5cm}{!}{
  \begin{tabular}{|c |  c  c c c c|} 
 \hline
 Source $\rightarrow$ Target  &C, P, R $\rightarrow$ A &A, P, R $\rightarrow$ C & C, A, R $\rightarrow$ P & C, P, A $\rightarrow$ R  &Ave. \\ [0.5ex] 
 %\hline\hline
  \hline
  
  Source only &54.3 &44.7 &69.3 &70.8 &59.8 \\

 % Source only &66.29 &81.25 &51.61 &79.25 &69.60  \\
  
 % CIDDG \cite{Li_2018_ECCV} &67.56 &82.87 &53.18 &80.69 &71.07 \\
  
 % CIDG \cite{DBLP:journals/corr/abs-1807-08479} &67.85 &82.59 &52.91 &80.37 &70.93 \\
  
 % DBADG \cite{8237853} &66.43 &81.51 &51.38 &79.82 &69.79 \\

DBADG \cite{8237853}   &54.8 &45.3 &70.3 & 70.6& 60.3\\

CIDG \cite{DBLP:journals/corr/abs-1807-08479}  &55.1 &45.8 &70.2 &71.4 &60.6\\
  
D-SAM &58.0 &44.4 &69.2 &71.5 &60.8 \\

CIDDG \cite{Li_2018_ECCV} &55.3 &46.2 &70.9 &71.9 &61.1\\

 JiGen \cite{carlucci2019domain} &53.0 &\textbf{47.5} &71.5 &72.8 &61.2 \\

 % Metareg &87.2 &79.2 &97.6 &70.3 \\
  \hline
 % \textbf{FNN(Ours)} &86.71 &78.49 &95.38 &74.94 &83.88 \\
  
   % \textbf{CFNN(Ours)} &\textbf{88.26} &\textbf{81.10} &\textbf{96.40} &\textbf{76.55} &\textbf{85.59}  \\
  \textbf{DDIAN (Ours)} &\textbf{57.9} &47.2 &\textbf{72.3} &\textbf{73.8} &\textbf{62.8} \\
  
 \hline
 \end{tabular}}
  \caption{Recognition accuracies for DG on the Office-Home dataset using pretrained ResNet-18.}
\label{table_off-home_ResNet}
\end{table*}

\begin{table*}[h!]
\fontsize{7}{7}\selectfont  
\centering
\small\addtolength{\tabcolsep}{11.5pt}
 %\begin{tabular}{||c c c c||} 
  \resizebox{14.5cm}{!}{
  \begin{tabular}{|c |  c  c c c c|} 
 \hline
 Source $\rightarrow$ Target  &P, C, S $\rightarrow$ A &P, A, S $\rightarrow$ C & A, C, S $\rightarrow$ P & A, C, P $\rightarrow$ S  &Ave. \\ [0.5ex] 
 %\hline\hline
  \hline
  
  Source only &77.8 &74.3 &94.7 &69.1 &79.0 \\ 
  
DDIAN (Global Domain Alignment)  &81.4 &74.2 &94.5 &69.1 &79.8 \\

DDIAN (Local Domain Alignment)  &79.6 &74.0 &94.6 &70.8 &79.8 \\

DDIAN (Discriminative Feature)  &80.2 &75.5 &94.9 &71.3 &80.5 \\ 
  
  \textbf{DDIAN (Ours)} &\textbf{83.4} &\textbf{76.7} &\textbf{95.3} &\textbf{74.1} &\textbf{82.4} \\
  
 % \hline
 %Ave.  & & & & &\\ [1ex] 
 
 \hline
 \end{tabular}}
  \caption{Recognition accuracies for DG on the PACS dataset \cite{8237853} using pretrained ResNet-18.}
\label{table_ablation_study}
\end{table*}

In this section, we demonstrate the experiments we have conducted to evaluate our proposed approach and compare the proposed approach with state-of-the-art domain generalization methods. 

\subsection{Datasets}
The proposed approach is evaluated on \textbf{PACS} \cite{8237853}, \textbf{Office-Home}  \cite{venkateswara2017Deep} and VLCS benchmarks in the context of domain generalization. 

\subsubsection{PACS}
The \textbf{PACS} \cite{8237853} domain generalization dataset is built by taking the common categories among Caltech256 , Sketchy, TU-Berlin and Google Images. It has 4 domains: Photo, Sketch, Cartoon and Painting. Each domain consists of 7 categories: dog, guitar, giraffe, elephant, person, horse, house. It contains total 9991 images. We evaluate our proposed method on four transfer tasks P, C, S $\rightarrow$ A; P, A, S $\rightarrow$ C; A, C, S $\rightarrow$ P; and A, C, P $\rightarrow$ P. The transfer task P, C, S $\rightarrow$ A indicates three source domains Photo (P), Cartoon (C) and Sketch (S) and one target domain Art-Painting (A). We follow the standard protocol for domain generalization where during the training phase we access the labeled source data but not access the target data. The target data is used only in test phase only.

\subsubsection{VLCS} 
VLCS is another cross-domain object benchmark that consists of the images from four popular datasets: PASCAL VOC2007 (V),
LabelMe (L), Caltech-101 (C), and SUN09 (S). The are five common classes, i.e., 'bird', 'dog', 'car', 'chair' and 'person' across four domains. We follow the same setting in where each domain of VLCS is divided into a training set (70\%) and a test set (30\%) trough random selection. We evaluate our proposed method on four transfer tasks L, P, S $\rightarrow$ C; P, C, S $\rightarrow$ L; C, L, S $\rightarrow$ P; and P, L, C $\rightarrow$ S. The transfer task L, P, S $\rightarrow$ C indicates there are three source domains LabelMe(L), PASCAL(P) and Sun09(S); and one target domain Caltech-101(C).

\subsubsection{Office-Home}
Office-Home dataset contains four domains named Art (Ar), Real-World (Rw), Clipart (Cl) and Product (Pr) with 65 different object categories. It has around 15,500 images with 65 categories. To build the Art and Real-world domains, public domain images were collected from websites such as www.deviantart.com and www.flickr.com. Clipart images were taken from various clipart webpages. The Product domain images were obtained utilizing web-crawlers from www.amazon.com. We evaluate our proposed method on four transfer tasks C, P, R $\rightarrow$ A; A, P, R $\rightarrow$ C; C, A, R $\rightarrow$ P; and C, P, A $\rightarrow$ R. The transfer task C, P, R $\rightarrow$ A indicates there are three source domains Clipart (C), Product (P) and Real-world (R); and one target domain Art (A).

\subsection{Comparison with state-of-the-art}

We compare the performance of DDIAN against several recent domain generalization methods. \textbf{Source only} is the simple source domains aggregation approach for DG without any adaptive loss function. \textbf{CIDDG} \cite{Li_2018_ECCV} is a deep DG method based on adversarial networks where the discrepancy among source domains is minimized by using class prior normalized domain classification and class conditional domain classification loss. \textbf{MLDG} \cite{li2018learning} is a meta-learning based DG method. \textbf{CIDG} \cite{DBLP:journals/corr/abs-1807-08479} is DG framework where both marginal and conditional representations are considered to mitigate the DA problem. \textbf{DBADG} \cite{8237853} is a DG framework based on low rank parameterized convolutional neural network. \textbf{CrossGrad} \cite{shankar2018generalizing} is a recent approach of disrupting the input manifold for DG utilising Bayesian networks. \textbf{MetaReg} \cite{balaji2018metareg} is a recently proposed approach for DG that meta-learns the classifier regularizer. \textbf{MMAL}\cite{finn2017model} is originally designed for domain adaptation and re-purposed for domain generalization. \textbf{MMD-AAE}\cite{Li_2018_CVPR} is a recent domain generalization approach that learns domain invariant features by aligning the features by MMD constraint. \textbf{CCSA}\cite{motiian2017CCSA} uses semantic alignment to regularize the learned feature subspace. \textbf{DSN}\cite{Bousmalis:2016:DSN:3157096.3157135} gains domain alignment by decomposing the source domains into private and shared spaces and learned them by reconstruction signal. \textbf{D-SAM}\cite{d2018domain} is a domain generalization approach based on the utilization of domain-specific aggregation modules. \textbf{MASF}\cite{dou2019domain} regularizes the semantic features by a gradient-based meta-training procedure. \textbf{EPI-FCR} \cite{song2019episodic} achieves domain invariant features for domain generalization by episodic training which is based on the domain aggregation method.

\subsection{Implementation Details}

We implement our proposed method based on the PyTorch deep learning framework. We fine-tune the network using either AlexNet or ResNet-18 models pretrained on the ImageNet dataset. All the convolutional and pooling layers are fine-tuned during the training and the classifier layer is trained from scratch by backpropagation for all the transfer tasks. We set the learning rate of the classifier to be 10 times compared to other layers as it is trained from scratch. We use mini-batch Stochastic Gradient Descent (SGD) with momentum of 0.9 for optimization and we change the learning rate as \cite{pmlr-v37-ganin15}. We set $\alpha = 1 $, $\beta = 0.5$, $\gamma =0.5$ , batch size = 32. We follow standard evaluation for domain generalization and use all source examples with labels during training. It is noted that the target data is unavailable during training. To compute the average accuracy, the results are obtained by running each transfer task 5 times.

\subsection{Results}

The classification accuracy on the four transfer tasks on the PACS dataset is reported in Table \ref{table_pacs_AlexNet} using pre-trained AlxNet architecture on ImageNet. From Table \ref{table_pacs_AlexNet}, we can see that our discriminative domain-invariant approach achieved comparable results on each transfer task and our proposed approach outperforms most of the state-of-the-art approaches except MASF and JiGen methods. We also observe that DDIAN provides the highest overall efficiency, with just 6\% progress on source only approach.

The classification performance on the four transfer tasks on PACS dataset using ResNet-18 is reported in Table 
\ref{table_pacs_ResNet}, we can observe that with ResNet-18 architecture, the obtained results are enhanced as anticipated across the board.  Our system nevertheless retains the highest overall efficiency, with a 3.4\ percent increase on source only approach.

Table \ref{vlcs} presents the classification performance on the four transfer tasks on VLCS dataset in the context of domain generalization. For VLCS dataset, we follow the same protocol of \cite{song2019episodic} where each domain is split into train (70\%) and test (30\%) and do leave-one-out evaluation. From the results we observe that our method outperforms prior state-of-the-art approaches for domain generalization. We obtained 6.4\% improvements over source only method in the four transfer domain generalization tasks.

The classification accuracy on the four transfer tasks on Office-Home dataset is shown in Table \ref{table_off-home_ResNet}. From the results, we can see that our proposed approach has obtained the highest results on the P, C, S $\rightarrow$ A; C, A, R $\rightarrow$P and C, P, A $\rightarrow$ R transfer tasks and comparable performance on A, P, R $\rightarrow$ C transfer task. DDIAN also provides the best performance overall, with 3\% improvement on source only approach, and at least 1.6 \% improvement on prior state-of-the-art methods CIDDG, D-SAM and JiGen.

\subsection{Ablation study}

In this section, we further conduct additional experiments using PACS dataset to investigate the contribution of each component of our proposed domain generalization network's performance. The average accuracies over five runs using individual components are shown in Table \ref{table_ablation_study}. We conduct experiments of three individual components: DDIAN (Global Domain Alignment), DDIAN (Local Domain Alignment) and DDIAN (Discriminative Feature). We remove two components and keep one component, DDIAN (Global Domain Alignment) means that we remove local domain alignment and discriminative feature components to see the contribution of global domain alignment component.  DDIAN (Local Domain Alignment) indicates that we remove global domain alignment and discriminative feature whereas DDIAN (Discriminative Feature) idicates that we remove both global and local domain alignment modules and keep only discriminative feature component with our baseline. From the ablation study, we observe that DDIAN with each individual component outperforms source only approach. It is also noted that DDIAN with all components outperform other approaches. From the results of the ablation study, we conclude that each component plays its own role for achieving domain-invariant features and gains generalization on unseen target data. We also found that the discriminative features are useful for generalization on the unknown target domain.

\section{Conclusion}
In this paper, we addressed the domain generalization issue where the discriminative features are extracted in an adversarial way. The adversarial module not only aligns the marginal distributions of the source domains but also aligns the conditional distributions of the domains. The approach of learning a discriminative feature is introduced to apply the feature space for better inter-class separability and intra-class compactness that can help both predicting the categories and domain compatibility. There are two factors that can contribute the deep-feature discriminativeness lead to achieve the domain agnostic model. On one side, since the deep representations are better clustered, it is much easier to perform the domain alignment. On the other side, there is a wide distance across the hyperplane and every cluster owing to the improved inter-class separability. Therefore, it is less probable to misclassify the samples scattered far from the middle of each cluster within a domain or close to the edge. We have demonstrated the effectiveness of the proposed approach on several benchmarks and have achieved the state-of-the-art performance in most of the transfer tasks. 

%\vspace{-2mm}

%\vspace{-4mm}
\section*{Acknowledgements}
%\vspace{-2mm}
The research presented in this paper was supported by
Australian Research Council (ARC) Discovery Project Grant DP170100632.

\bibliographystyle{unsrt}  
\bibliography{references}  %%% Remove comment to use the external .bib file (using bibtex).

\begin{thebibliography}{10}

\bibitem{chen2020adversarial}
Minghao Chen, Shuai Zhao, Haifeng Liu, and Deng Cai.
\newblock Adversarial-learned loss for domain adaptation.
\newblock In {\em Proceedings of the AAAI Conference on Artificial
  Intelligence}, volume~34, pages 3521--3528, 2020.

\bibitem{lee2019drop}
Seungmin Lee, Dongwan Kim, Namil Kim, and Seong-Gyun Jeong.
\newblock Drop to adapt: Learning discriminative features for unsupervised
  domain adaptation.
\newblock In {\em Proceedings of the IEEE International Conference on Computer
  Vision}, pages 91--100, 2019.

\bibitem{Rahman2020}
Mohammad~Mahfujur Rahman, Clinton Fookes, Mahsa Baktashmotlagh, and Sridha
  Sridharan.
\newblock {\em On Minimum Discrepancy Estimation for Deep Domain Adaptation},
  pages 81--94.
\newblock Springer International Publishing, Cham, 2020.

\bibitem{liu2019transferable}
Hong Liu, Mingsheng Long, Jianmin Wang, and Michael Jordan.
\newblock Transferable adversarial training: A general approach to adapting
  deep classifiers.
\newblock In {\em International Conference on Machine Learning}, pages
  4013--4022, 2019.

\bibitem{ma2019gcan}
Xinhong Ma, Tianzhu Zhang, and Changsheng Xu.
\newblock Gcan: Graph convolutional adversarial network for unsupervised domain
  adaptation.
\newblock In {\em Proceedings of the IEEE Conference on Computer Vision and
  Pattern Recognition}, pages 8266--8276, 2019.

\bibitem{zhang2020unsupervised}
Yabin Zhang, Bin Deng, Hui Tang, Lei Zhang, and Kui Jia.
\newblock Unsupervised multi-class domain adaptation: Theory, algorithms, and
  practice.
\newblock {\em IEEE Transactions on Pattern Analysis and Machine Intelligence},
  2020.

\bibitem{mahfujur2021preserving}
Mohammad Mahfujur~Rahman, Clinton Fookes, and Sridha Sridharan.
\newblock Preserving semantic consistency in unsupervised domain adaptation
  using generative adversarial networks.
\newblock {\em arXiv e-prints}, pages arXiv--2104, 2021.

\bibitem{qiu2020partial}
Wenjie Qiu, Wendong Chen, and Haifeng Hu.
\newblock Partial domain adaptation based on shared class oriented adversarial
  network.
\newblock {\em Computer Vision and Image Understanding}, 199:103018, 2020.

\bibitem{su2021multi}
Hang Su, Shaogang Gong, and Xiatian Zhu.
\newblock Multi-perspective cross-class domain adaptation for open logo
  detection.
\newblock {\em Computer Vision and Image Understanding}, 204:103156, 2021.

\bibitem{du2020learning}
Yingjun Du, Jun Xu, Huan Xiong, Qiang Qiu, Xiantong Zhen, Cees~GM Snoek, and
  Ling Shao.
\newblock Learning to learn with variational information bottleneck for domain
  generalization.
\newblock In {\em European Conference on Computer Vision}, pages 200--216.
  Springer, 2020.

\bibitem{rahman2019multi}
Mohammad~Mahfujur Rahman, Clinton Fookes, Mahsa Baktashmotlagh, and Sridha
  Sridharan.
\newblock Multi-component image translation for deep domain generalization.
\newblock In {\em 2019 IEEE Winter Conference on Applications of Computer
  Vision (WACV)}, pages 579--588. IEEE, 2019.

\bibitem{RAHMAN2020107124}
Mohammad~Mahfujur Rahman, Clinton Fookes, Mahsa Baktashmotlagh, and Sridha
  Sridharan.
\newblock Correlation-aware adversarial domain adaptation and generalization.
\newblock {\em Pattern Recognition}, 100, 2020.

\bibitem{li2018domain}
Haoliang Li, Sinno Jialin~Pan, Shiqi Wang, and Alex~C Kot.
\newblock Domain generalization with adversarial feature learning.
\newblock In {\em Proceedings of the IEEE Conference on Computer Vision and
  Pattern Recognition}, pages 5400--5409, 2018.

\bibitem{qiao2020learning}
Fengchun Qiao, Long Zhao, and Xi~Peng.
\newblock Learning to learn single domain generalization.
\newblock In {\em Proceedings of the IEEE/CVF Conference on Computer Vision and
  Pattern Recognition}, pages 12556--12565, 2020.

\bibitem{ryu2019generalized}
Jongbin Ryu, Gitaek Kwon, Ming-Hsuan Yang, and Jongwoo Lim.
\newblock Generalized convolutional forest networks for domain generalization
  and visual recognition.
\newblock In {\em International Conference on Learning Representations}, 2019.

\bibitem{zhou2020learning}
Kaiyang Zhou, Yongxin Yang, Timothy Hospedales, and Tao Xiang.
\newblock Learning to generate novel domains for domain generalization.
\newblock In {\em European Conference on Computer Vision}, pages 561--578.
  Springer, 2020.

\bibitem{tseng2020cross}
Hung-Yu Tseng, Hsin-Ying Lee, Jia-Bin Huang, and Ming-Hsuan Yang.
\newblock Cross-domain few-shot classification via learned feature-wise
  transformation.
\newblock {\em arXiv preprint arXiv:2001.08735}, 2020.

\bibitem{matsuura2020domain}
Toshihiko Matsuura and Tatsuya Harada.
\newblock Domain generalization using a mixture of multiple latent domains.
\newblock In {\em Proceedings of the AAAI Conference on Artificial
  Intelligence}, volume~34, pages 11749--11756, 2020.

\bibitem{chattopadhyay2020learning}
Prithvijit Chattopadhyay, Yogesh Balaji, and Judy Hoffman.
\newblock Learning to balance specificity and invariance for in and out of
  domain generalization.
\newblock In {\em European Conference on Computer Vision}, pages 301--318.
  Springer, 2020.

\bibitem{seo2019learning}
Seonguk Seo, Yumin Suh, Dongwan Kim, Jongwoo Han, and Bohyung Han.
\newblock Learning to optimize domain specific normalization for domain
  generalization.
\newblock {\em arXiv preprint arXiv:1907.04275}, 3(6):7, 2019.

\bibitem{wang2020learning}
Shujun Wang, Lequan Yu, Caizi Li, Chi-Wing Fu, and Pheng-Ann Heng.
\newblock Learning from extrinsic and intrinsic supervisions for domain
  generalization.
\newblock In {\em European Conference on Computer Vision}, pages 159--176.
  Springer, 2020.

\bibitem{mahfujur2021deep}
Mohammad Mahfujur~Rahman, Clinton Fookes, and Sridha Sridharan.
\newblock Deep domain generalization with feature-norm network.
\newblock {\em arXiv e-prints}, pages arXiv--2104, 2021.

\bibitem{Li2018eccv}
Ya~Li, Xinmei Tian, Mingming Gong, Yajing Liu, Tongliang Liu, Kun Zhang, and
  Dacheng Tao.
\newblock Deep domain generalization via conditional invariant adversarial
  networks.
\newblock In {\em ECCV}, 2018.

\bibitem{Li_2018_CVPR}
Haoliang Li, Sinno Jialin~Pan, Shiqi Wang, and Alex~C. Kot.
\newblock Domain generalization with adversarial feature learning.
\newblock In {\em CVPR}, 2018.

\bibitem{blanchard2017domain}
Gilles Blanchard, Aniket~Anand Deshmukh, Urun Dogan, Gyemin Lee, and Clayton
  Scott.
\newblock Domain generalization by marginal transfer learning.
\newblock {\em arXiv preprint arXiv:1711.07910}, 2017.

\bibitem{NIPS2011_4312}
Gilles Blanchard, Gyemin Lee, and Clayton Scott.
\newblock Generalizing from several related classification tasks to a new
  unlabeled sample.
\newblock In {\em Advances in Neural Information Processing Systems (NIPS)}.
  2011.

\bibitem{Muandet:2013:DGV:3042817.3042820}
Krikamol Muandet, David Balduzzi, and Bernhard Sch\"{o}lkopf.
\newblock Domain generalization via invariant feature representation.
\newblock In {\em Proceedings of the International Conference on International
  Conference on Machine Learning (ICML)}, 2013.

\bibitem{DBLP:journals/pami/GhifaryBKZ17}
Muhammad Ghifary, David Balduzzi, W.~Bastiaan Kleijn, and Mengjie Zhang.
\newblock Scatter component analysis: {A} unified framework for domain
  adaptation and domain generalization.
\newblock {\em TPAMI}, 39(7), 2017.

\bibitem{ECCV12_Khosla}
Aditya Khosla, Tinghui Zhou, Tomasz Malisiewicz, Alexei Efros, and Antonio
  Torralba.
\newblock Undoing the damage of dataset bias.
\newblock In {\em European Conference on Computer Vision (ECCV)}, Florence,
  Italy, October 2012.

\bibitem{DBLP:conf/iccv/GhifaryKZB15}
Muhammad Ghifary, W.~Bastiaan Kleijn, Mengjie Zhang, and David Balduzzi.
\newblock Domain generalization for object recognition with multi-task
  autoencoders.
\newblock In {\em IEEE International Conference on Computer Vision, (ICCV)},
  2015.

\bibitem{Xu2014}
Zheng Xu, Wen Li, Li~Niu, and Dong Xu.
\newblock {\em Exploiting Low-Rank Structure from Latent Domains for Domain
  Generalization}, pages 628--643.
\newblock Springer International Publishing, Cham, 2014.

\bibitem{FXR_iccv13}
Chen Fang, Ye~Xu, and Daniel~N. Rockmore.
\newblock Unbiased metric learning: On the utilization of multiple datasets and
  web images for softening bias.
\newblock In {\em International Conference on Computer Vision}, 2013.

\bibitem{7733141}
L.~Niu, W.~Li, D.~Xu, and J.~Cai.
\newblock An exemplar-based multi-view domain generalization framework for
  visual recognition.
\newblock {\em IEEE Transactions on Neural Networks and Learning Systems},
  29(2), 2018.

\bibitem{7298894}
L.~Niu, W.~Li, and D.~Xu.
\newblock Visual recognition by learning from web data: A weakly supervised
  domain generalization approach.
\newblock In {\em IEEE Conference on Computer Vision and Pattern Recognition
  (CVPR)}, pages 2774--2783, June 2015.

\bibitem{volpi2018generalizing}
Riccardo Volpi, Hongseok Namkoong, Ozan Sener, John~C Duchi, Vittorio Murino,
  and Silvio Savarese.
\newblock Generalizing to unseen domains via adversarial data augmentation.
\newblock In {\em Advances in Neural Information Processing Systems}, pages
  5334--5344, 2018.

\bibitem{8237853}
D.~Li, Y.~Yang, Y.~Z. Song, and T.~M. Hospedales.
\newblock Deeper, broader and artier domain generalization.
\newblock In {\em IEEE International Conference on Computer Vision (ICCV)},
  2017.

\bibitem{song2019episodic}
Yi-Zhe Song.
\newblock Episodic training for domain generalization.
\newblock In {\em Proceedings of the International Conference on Computer
  Vision (ICCV 2019)}. Institute of Electrical and Electronics Engineers
  (IEEE), 2019.

\bibitem{balaji2018metareg}
Yogesh Balaji, Swami Sankaranarayanan, and Rama Chellappa.
\newblock Metareg: Towards domain generalization using meta-regularization.
\newblock In {\em Advances in Neural Information Processing Systems}, pages
  998--1008, 2018.

\bibitem{li2019feature}
Yiying Li, Yongxin Yang, Wei Zhou, and Timothy~M Hospedales.
\newblock Feature-critic networks for heterogeneous domain generalization.
\newblock {\em arXiv preprint arXiv:1901.11448}, 2019.

\bibitem{carlucci2019domain}
Fabio~M Carlucci, Antonio D'Innocente, Silvia Bucci, Barbara Caputo, and
  Tatiana Tommasi.
\newblock Domain generalization by solving jigsaw puzzles.
\newblock In {\em Proceedings of the IEEE Conference on Computer Vision and
  Pattern Recognition}, pages 2229--2238, 2019.

\bibitem{shankar2018generalizing}
Shiv Shankar, Vihari Piratla, Soumen Chakrabarti, Siddhartha Chaudhuri, Preethi
  Jyothi, and Sunita Sarawagi.
\newblock Generalizing across domains via cross-gradient training.
\newblock {\em arXiv preprint arXiv:1804.10745}, 2018.

\bibitem{asadi2019towards}
Nader Asadi, Mehrdad Hosseinzadeh, and Mahdi Eftekhari.
\newblock Towards shape biased unsupervised representation learning for domain
  generalization.
\newblock {\em arXiv preprint arXiv:1909.08245}, 2019.

\bibitem{deshmukh2019generalization}
Aniket~Anand Deshmukh, Yunwen Lei, Srinagesh Sharma, Urun Dogan, James~W
  Cutler, and Clayton Scott.
\newblock A generalization error bound for multi-class domain generalization.
\newblock {\em arXiv preprint arXiv:1905.10392}, 2019.

\bibitem{akuzawa2019adversarial}
Kei Akuzawa, Yusuke Iwasawa, and Yutaka Matsuo.
\newblock Adversarial invariant feature learning with accuracy constraint for
  domain generalization.
\newblock {\em arXiv preprint arXiv:1904.12543}, 2019.

\bibitem{truong2018beyond}
Thanh-Dat Truong, Chi~Nhan Duong, Khoa Luu, Minh-Triet Tran, and Minh Do.
\newblock Beyond domain adaptation: Unseen domain encapsulation via universal
  non-volume preserving models.
\newblock {\em arXiv preprint arXiv:1812.03407}, 2018.

\bibitem{ilse2019diva}
Maximilian Ilse, Jakub~M Tomczak, Christos Louizos, and Max Welling.
\newblock Diva: Domain invariant variational autoencoders.
\newblock {\em arXiv preprint arXiv:1905.10427}, 2019.

\bibitem{dou2019domain}
Qi~Dou, Daniel~Coelho de~Castro, Konstantinos Kamnitsas, and Ben Glocker.
\newblock Domain generalization via model-agnostic learning of semantic
  features.
\newblock In {\em Advances in Neural Information Processing Systems}, pages
  6447--6458, 2019.

\bibitem{8053784}
Z.~Ding and Y.~Fu.
\newblock Deep domain generalization with structured low-rank constraint.
\newblock {\em TIP}, 27(1), 2018.

\bibitem{motiian2017CCSA}
Saeid Motiian, Marco Piccirilli, Donald~A. Adjeroh, and Gianfranco Doretto.
\newblock Unified deep supervised domain adaptation and generalization.
\newblock In {\em IEEE International Conference on Computer Vision (ICCV)},
  2017.

\bibitem{wang2019learning}
Haohan Wang, Zexue He, Zachary~C Lipton, and Eric~P Xing.
\newblock Learning robust representations by projecting superficial statistics
  out.
\newblock {\em arXiv preprint arXiv:1903.06256}, 2019.

\bibitem{DBLP:journals/corr/abs-1807-08479}
Ya~Li, Mingming Gong, Xinmei Tian, Tongliang Liu, and Dacheng Tao.
\newblock Domain generalization via conditional invariant representation.
\newblock In {\em AAAI}, 2018.

\bibitem{li2018deep}
Ya~Li, Xinmei Tian, Mingming Gong, Yajing Liu, Tongliang Liu, Kun Zhang, and
  Dacheng Tao.
\newblock Deep domain generalization via conditional invariant adversarial
  networks.
\newblock In {\em Proceedings of the European Conference on Computer Vision
  (ECCV)}, pages 624--639, 2018.

\bibitem{matsuura2019domain}
Toshihiko Matsuura and Tatsuya Harada.
\newblock Domain generalization using a mixture of multiple latent domains.
\newblock {\em arXiv preprint arXiv:1911.07661}, 2019.

\bibitem{Ganin:2016:DTN:2946645.2946704}
Yaroslav Ganin, Evgeniya Ustinova, Hana Ajakan, Pascal Germain, Hugo
  Larochelle, Fran\c{c}ois Laviolette, Mario Marchand, and Victor Lempitsky.
\newblock Domain-adversarial training of neural networks.
\newblock {\em JMLR}, 17(1), 2016.

\bibitem{khatun2021end}
Amena Khatun, Simon Denman, Sridha Sridharan, and Clinton Fookes.
\newblock End-to-end domain adaptive attention network for cross-domain person
  re-identification.
\newblock {\em IEEE Transactions on Information Forensics and Security}, 2021.

\bibitem{khatun2020joint}
Amena Khatun, Simon Denman, Sridha Sridharan, and Clinton Fookes.
\newblock Joint identification--verification for person re-identification: A
  four stream deep learning approach with improved quartet loss function.
\newblock {\em Computer Vision and Image Understanding}, 197:102989, 2020.

\bibitem{khatun2018deep}
Amena Khatun, Simon Denman, Sridha Sridharan, and Clinton Fookes.
\newblock A deep four-stream siamese convolutional neural network with joint
  verification and identification loss for person re-detection.
\newblock In {\em 2018 IEEE Winter Conference on Applications of Computer
  Vision (WACV)}, pages 1292--1301. IEEE, 2018.

\bibitem{khatun2020semantic}
Amena Khatun, Simon Denman, Sridha Sridharan, and Clinton Fookes.
\newblock Semantic consistency and identity mapping multi-component generative
  adversarial network for person re-identification.
\newblock In {\em Proceedings of the IEEE/CVF Winter Conference on Applications
  of Computer Vision}, pages 2267--2276, 2020.

\bibitem{wen2016discriminative}
Yandong Wen, Kaipeng Zhang, Zhifeng Li, and Yu~Qiao.
\newblock A discriminative feature learning approach for deep face recognition.
\newblock In {\em European conference on computer vision}, pages 499--515.
  Springer, 2016.

\bibitem{Li_2018_ECCV}
Ya~Li, Xinmei Tian, Mingming Gong, Yajing Liu, Tongliang Liu, Kun Zhang, and
  Dacheng Tao.
\newblock Deep domain generalization via conditional invariant adversarial
  networks.
\newblock In {\em ECCV}, 2018.

\bibitem{NIPS2016_6254}
Konstantinos Bousmalis, George Trigeorgis, Nathan Silberman, Dilip Krishnan,
  and Dumitru Erhan.
\newblock Domain separation networks.
\newblock In {\em NIPS}. 2016.

\bibitem{li2018learning}
Da~Li, Yongxin Yang, Yi-Zhe Song, and Timothy~M. Hospedales.
\newblock Learning to generalize: Meta-learning for domain generalization.
\newblock In {\em AAAI}, 2018.

\bibitem{d2018domain}
Antonio D’Innocente and Barbara Caputo.
\newblock Domain generalization with domain-specific aggregation modules.
\newblock In {\em German Conference on Pattern Recognition}, pages 187--198.
  Springer, 2018.

\bibitem{finn2017model}
Chelsea Finn, Pieter Abbeel, and Sergey Levine.
\newblock Model-agnostic meta-learning for fast adaptation of deep networks.
\newblock In {\em Proceedings of the 34th International Conference on Machine
  Learning}, 2017.

\bibitem{venkateswara2017Deep}
Venkateswara Hemanth, Eusebio Jose, Chakraborty Shayok, and Panchanathan
  Sethuraman.
\newblock Deep hashing network for unsupervised domain adaptation.
\newblock In {\em IEEE Conference on Computer Vision and Pattern Recognition
  (CVPR)}, 2017.

\bibitem{Bousmalis:2016:DSN:3157096.3157135}
Konstantinos Bousmalis, George Trigeorgis, Nathan Silberman, Dilip Krishnan,
  and Dumitru Erhan.
\newblock Domain separation networks.
\newblock In {\em Proceedings of the International Conference on Neural
  Information Processing Systems (NIPS)}, 2016.

\bibitem{pmlr-v37-ganin15}
Yaroslav Ganin and Victor Lempitsky.
\newblock Unsupervised domain adaptation by backpropagation.
\newblock In {\em ICML}, 2015.

\end{thebibliography}

%\begin{thebibliography}{1}

%\bibitem{kour2014real}
%George Kour and Raid Saabne.
%\newblock Real-time segmentation of on-line handwritten arabic script.
%\newblock In {\em Frontiers in Handwriting Recognition (ICFHR), 2014 14th
 % International Conference on}, pages 417--422. IEEE, 2014.

%\bibitem{kour2014fast}
%George Kour and Raid Saabne.
%\newblock Fast classification of handwritten on-line arabic characters.
%\newblock In {\em Soft Computing and Pattern Recognition (SoCPaR), 2014 6th
  %International Conference of}, pages 312--318. IEEE, 2014.

%\bibitem{hadash2018estimate}
%Guy Hadash, Einat Kermany, Boaz Carmeli, Ofer Lavi, George Kour, and Alon
 % Jacovi.
%\newblock Estimate and replace: A novel approach to integrating deep neural
 % networks with existing applications.
%\newblock {\em arXiv preprint arXiv:1804.09028}, 2018.

%\end{thebibliography}

\end{document}